\documentclass[conference]{IEEEtran}
\usepackage{amsmath,amssymb}
\usepackage{booktabs}
\usepackage{graphicx}
\usepackage{multirow}
\usepackage{url}
\usepackage[table]{xcolor}
\usepackage{balance}

\newcommand{\Rsq}{R^2}

\begin{document}

\title{An Embedded RISC-V Evaluation of
Kolmogorov--Arnold Networks in Hard-Constrained Recurrent
Physics-Informed Models}

\author{
\IEEEauthorblockN{Enzo Nicol\'as Spotorno, Josafat Leal Filho}
\IEEEauthorblockA{Department of Informatics and Statistics - Graduate Program in Computer Science (PPGCC)\\
Federal University of Santa Catarina, Florian\'opolis, Brazil\\
\{enzoniko, josafat\}@posgrad.ufsc.br}
}

\maketitle

\begin{abstract}
Hard-constrained recurrent physics-informed networks (HRPINNs) embed known
dynamics inside a recurrent numerical integrator and restrict a neural branch
to learning only the residual dynamics that the first-principles model does
not capture. Kolmogorov--Arnold Networks (KANs) have been proposed as
parameter-efficient replacements for multilayer perceptrons (MLPs) in such
residual branches, but their learnable B-spline activations follow a markedly
different execution profile. Building on prior work that characterized when a
vanilla B-spline KAN matches or underperforms an MLP as an HRPINN residual
branch in discovery accuracy, this paper asks whether that parameter
efficiency survives deployment. Using identical trained weights, we measured
execution latency, energy per integration step, and dependability under
post-training quantization in the closed recurrent loop on a RISC-V RV64GC platform without vector extensions (StarFive VisionFive~2, SiFive U74). For
the two accuracy-comparable pairs, the KAN residual branch
executed $13.5\times$ and $8.0\times$ slower and consumed $11.3\times$ and
$5.6\times$ more energy per integration step (3.7\,$\mu$J against
0.33\,$\mu$J for the smallest pair); across all four parameter-matched size
tiers the ranges are $4.7\times$--$14.5\times$ and
$4.7\times$--$18.7\times$. Under INT8 quantization, KAN trajectories diverged up to $43\times$ earlier
than matched MLPs; the damage traces to weight quantization, not to
input-side knot-interval misassignment.
These results indicate that the parameter efficiency reported for KANs does
not transfer to deployment cost on scalar embedded cores, and that an MLP
residual branch is the more dependable default for embedded HRPINN
deployment unless specific quantization co-design is used.
\end{abstract}

\begin{IEEEkeywords}
Kolmogorov--Arnold networks, physics-informed learning, embedded systems,
RISC-V, quantization, dependability, cyber-physical systems
\end{IEEEkeywords}

\section{Introduction}
\label{sec:intro}

Modern cyber-physical systems (CPS) increasingly rely on executable models of
their own physics for prediction, anomaly detection, and control. When such a
model runs on the system itself, under real-time and energy budgets, its
predictions must also remain trustworthy, since a learned component that
degrades under deployment conditions can propagate into the control decisions
it informs~\cite{brunke2022safe}. Hard-constrained architectures address this
structurally rather than through training pressure, by embedding the imposed
constraints in the computation instead of in a loss term
(Section~\ref{sec:related}). The Hybrid Recurrent Physics-Informed Neural
Network (HRPINN)~\cite{spotorno2025hard} applies this principle to a
recurrent setting: the known dynamics and the numerical integrator are fixed
inside the update rule, and a neural branch learns only the residual dynamics
that the first-principles model does not capture, so the imposed structure
holds for any behavior of that branch, with similar approaches having also been studied across CPS prognostics, including
battery-health and remaining-useful-life
estimation~\cite{nascimento2021hybrid,rai2020hybrid}.

Because that residual branch is the only learned component, the architecture
chosen for it determines the model's deployment cost. Kolmogorov--Arnold
Networks (KANs)~\cite{liu2024kan} have been proposed as an alternative to
MLPs for scientific machine learning. They replace fixed node activations
with learnable univariate B-spline functions on each edge, summed at every
node, and have shown promise for recovering hidden terms in dynamical systems
(Section~\ref{sec:related}). In prior work~\cite{spotorno2026kaniclr},
summarized in Section~\ref{sec:background}, we characterized when a
\emph{vanilla} B-spline KAN matches or underperforms an MLP as an HRPINN
residual branch in discovery accuracy and training
stability. That characterization concerns accuracy, not execution
cost. Studies of KAN hardware implementation consistently report B-spline
evaluation as expensive relative to parameter
count~\cite{huang2025hardware,khalid2026hardware}, but they characterize KANs
as generic function approximators, without an enclosing physics-informed
integrator and without an accuracy figure attached to the specific model
measured. This leaves open the question that motivates this paper:
\textbf{does the parameter efficiency KANs exhibit on separable residuals
translate into computational efficiency once deployed, or does the B-spline
formulation introduce execution- and precision-level costs that offset it?}

We investigate three hypotheses. \textbf{H1}: the KAN residual branch incurs
higher per-step execution latency than an accuracy-comparable MLP.
\textbf{H2}: that latency difference translates into higher energy per
integration step. \textbf{H3}: the KAN branch loses trajectory fidelity
earlier than the MLP under post-training quantization inside the closed
recurrent loop, a regime in which memory and instruction-set constraints
often make integer execution the only option. Testing these hypotheses
requires a platform on which the cost of each computational topology is
exposed directly. We target the StarFive VisionFive~2 (SiFive U74,
RV64GC)~\cite{visionfive2}, which provides no vector extension, so both the
MLP's matrix products and the KAN's spline recursion compile to scalar
instruction sequences and neither can draw on vendor-tuned vector kernels;
the measured difference therefore reflects the operation mix and
memory-access pattern of the two topologies themselves.

Thus, this paper contributes: (i)~latency and energy measurements on a RISC-V
core without vector extensions, on identical trained weights, including a
deployment-optimized KAN variant that quantifies the sensitivity of the
comparison to implementation effort; (ii)~a dependability evaluation under
post-training quantization in the closed loop, with an attribution analysis
identifying which quantization step drives the difference between
architectures; and (iii)~deployment guidance for choosing a residual-branch
architecture in embedded physics-informed CPS. The inference engine, the
trained models, and the measurement scripts are publicly available.\footnote{All resources are publicly available at \url{https://github.com/enzoniko/khrpinn}} 

\section{Related Work}
\label{sec:related}

Physics-informed neural networks impose physical laws as a loss
penalty~\cite{raissi2019pinn}, which leaves constraint satisfaction
unguaranteed at inference; architecture-level enforcement makes the
constraint structural instead, for example through projection layers for
linear equality constraints~\cite{chen2024hard}, and
HRPINN~\cite{spotorno2025hard} fixes both the known dynamics and the
integrator inside a recurrent cell, the broader motivation for such
guarantees being surveyed by Brunke et al.~\cite{brunke2022safe}. Within that
setting, KANs have been applied to dynamical systems principally as
neural-ODE backbones recovering hidden physics, symbolic source terms and
structured latent states~\cite{liu2024kan,koenig2024kan,liu2025structured}.
The hyperparameter sensitivity documented in practical
guidance~\cite{noorizadegan2025practitioner} motivates the stability
characterization summarized in Section~\ref{sec:background}, and a benchmark
of physics-informed KANs independently reports that the original B-spline
parameterization lacks both accuracy and efficiency against MLP-based PINNs
and remains sensitive to random seed~\cite{shukla2024fair}, a record to which
this paper adds the deployment dimension. Many variants replace the B-spline
basis with Chebyshev, operator-network or hybrid recurrent
formulations~\cite{mostajeran2025scaled,abueidda2025deepokan,zhang2025physics},
but all retain the defining primitive of a per-edge univariate function
evaluation followed by summation, so this paper evaluates the baseline
vanilla formulation as the reference an optimized descendant would have to
beat. On the hardware side, Huang et al.~\cite{huang2025hardware} show that
mapping B-splines into lookup tables still demands substantial circuit
resources and propose an algorithm-hardware co-design, while Khalid et
al.~\cite{khalid2026hardware} derive platform-independent complexity formulae
across basis variants; both evaluate KANs as generic function approximators
on dedicated or abstract hardware, which is the gap this paper addresses.

\section{Background and Prior Results}
\label{sec:background}

HRPINN advances a state vector $s_t = [x_t, v_t]^{\top} \in \mathbb{R}^2$,
where $x_t$ is position and $v_t$ velocity in a normalized coordinate frame,
through a hard-constrained explicit-Euler cell:
$\dot v_t = a_{\mathrm{known}}(s_t) + R_\theta(s_t)$,
$x_{t+1} = x_t + \Delta t\, v_t$, and
$v_{t+1} = v_t + \Delta t\, \dot v_t$. Both $a_{\mathrm{known}}$ and
$R_\theta$ take the full state vector as argument; for the Van der Pol system
below the known term depends on $x_t$ alone while the residual depends on
both. The integrator is explicit (forward) Euler, as in the prior
study~\cite{spotorno2026kaniclr}, which keeps the trained models and their
reported accuracy directly transferable to this evaluation, and which is also
the cheapest choice, requiring one residual-branch evaluation per step
against $p$ for a Runge--Kutta scheme of order $p$. For the comparison
reported here the decisive property is that the integrator is identical and
fixed across both architectures: it contributes an additive constant to every
measurement and cancels in every architecture-to-architecture ratio. The learned branch is evaluated once per step and the remaining terms are
closed-form expressions of a few floating-point operations, so the per-step
cost is dominated by the branch; where the first-principles term is itself
expensive, requiring an implicit solve or a large tabulated lookup, the
costs reported here bound only the learned part of the budget.

Only $R_\theta$ is trained, by backpropagation through time (BPTT) over
20-step windows. Because $R_\theta$ is the only component that changes when
the residual-branch architecture changes, comparing two architectures under
this cell isolates the comparison to the branch itself. The benchmark systems
are the Duffing oscillator ($\dot v = -x - 0.2v - 0.3x^3$; known part
$-x - 0.2v$, residual $-0.3x^3$) and the Van der Pol oscillator
($\dot v = -x + (1-x^2)v$; known part $-x$, residual $(1-x^2)v$). Both have
unit natural frequency, giving a characteristic period
$T = 2\pi \approx 6.28$\,s, so the step $\Delta t = 0.05$\,s corresponds to
$T/126$, a resolution at which explicit Euler remains stable for these
dynamics, and the 20-step training window spans $1.0$\,s, or $T/6$, long
enough for compounding error to appear in the loss while keeping the unrolled
graph small enough for stable gradients. Expressing both as fractions of the
system timescale lets them transfer to plants with different natural
frequencies. Discovery accuracy is reported as the $\Rsq$ between the learned
surface $R_\theta(x,v)$ and the analytical residual on a $50\times50$ grid
over $[-2,2]^2$. The residual branch is implemented as a ReLU MLP or as a vanilla B-spline
KAN. A KAN layer replaces each fixed scalar activation of an MLP with a
learnable univariate function, implemented as a B-spline of order $k$ over
$G$ grid intervals, combining base and spline paths as
$\phi(x) = w_b\,\mathrm{SiLU}(x) + w_s\sum_j c_j B_j(x)$, where $B_j$ are the
order-$k$ basis functions on that edge's knot vector and $c_j$ are learned
coefficients~\cite{liu2024kan}. Each output node sums its incoming edges, so
a single layer combines its inputs additively, and a multiplicative
interaction such as the Van der Pol residual $(1-x^2)v$ can be represented
only through composition across depth; the Van der Pol benchmark probes
exactly this. The default configuration uses grid size $G=5$ and spline order
$k=3$, and training applies grid updates, so the knot vector is generally
non-uniform and specific to the trained model.

The prior study ran two ablations, each over 100 random seeds with 95\%
bootstrap confidence intervals. A configuration ablation over grid size,
spline-scale weighting, sparsity regularization and grid-update frequency,
for a fixed KAN architecture ($[2,8,1]$), found accuracy highly sensitive to
these hyperparameters: three of seven configurations produced a negative Van
der Pol $\Rsq$, the worst at $-5.23$ against $0.77$ for a 337-parameter MLP
reference, a sensitivity absent in the MLP and consistent with reported KAN
practice~\cite{noorizadegan2025practitioner}; the two most stable settings,
$G{=}5$ and $G{=}3$, are carried forward as the KAN defaults. A
parameter-efficiency ablation over four KAN and four MLP widths and depths
(Table~\ref{tab:results}) shows three behaviors that frame this evaluation:
small KANs match or exceed similarly sized MLPs on the separable, univariate
Duffing residual; KANs consistently fail to recover the multiplicative Van
der Pol interaction, where MLPs scale gracefully with parameter count; and
the Wide and Deep KAN configurations are unstable, failing to
converge in most seeds on Duffing.

\section{Embedded Inference Framework}
\label{sec:framework}

Training is performed off-line in PyTorch on an x86-64 development host; deployment proceeds through export to a compact binary, verification of numerical equivalence, and execution by a scalar inference engine that depends only on the C++17 standard library and libm and is linked statically, so deployment is a copy of two files onto a minimal Linux image. The benchmark executable occupies 893\,KB and the model files range from 511\,B (MLP Tiny) to 17.8\,KB (MLP Large), against several hundred megabytes for a PyTorch runtime, justifying not training on the target platform. The exported file is a flat, little-endian, positional binary: a fixed header (magic, version, architecture and system identifiers, $\Delta t$, state normalizer) followed by per-layer records holding, for an MLP, the weight matrix, bias vector and activation flag, and for a KAN, the knot vector, base weights and spline coefficients with the per-edge scale already folded in, so no scaling multiply remains in the inference path. Each record also carries the per-tensor calibration ranges used by the quantized path, so a quantized deployment needs no run-time calibration data, and weights are stored in the layout the kernels iterate over, so loading is a single contiguous read with no dynamic allocation. All models fit within the U74's 32\,KB L1 data cache, keeping the latency comparison a measure of computation rather than of memory-hierarchy behavior.

Both branches are allocation-free, contiguous scalar loops. The MLP is a
chain of fused matrix-vector product-and-ReLU passes over row-major weights;
the KAN evaluates, per input feature, the SiLU base path and the
Cox--de~Boor spline recursion over that feature's knot vector, followed by a
spline matrix-vector product. Two properties of the KAN kernel matter for the
measurements that follow. First, the SiLU base path requires one $e^{-x}$
evaluation per input feature, and transcendental functions have no single
hardware instruction on the cores evaluated here: libm resolves them through
iterative polynomial approximations (\texttt{expf}) costing several times a
multiply-accumulate, with a range-reduction step that branches on the input
magnitude, which is why Table~\ref{tab:results} counts them separately.
Second, the KAN's memory access is
data-dependent: the MLP's forward pass is a fixed sequence of matrix-vector
products over contiguous memory, friendly to prefetching, whereas the spline
recursion selects a different active knot interval, and therefore a different
offset into the spline-coefficient tensor, for every input value.

The engine implements the spline recursion in two variants: a default that
reproduces the reference PyTorch numerics operation for operation, including
explicit divisions by knot differences, which is what allows it to be pinned
against the training framework, and an optimized variant that precomputes the
reciprocal of every knot-difference denominator at model load, legitimate
because the knot vector is frozen after training. Both are reported, since
omitting the optimization would understate the KAN's achievable performance.
In every other respect the architectures receive identical treatment: the same compiler and flags (\texttt{-O2}, \texttt{-ffp-contract=off}), single-threaded execution, and the same measurement harness. The engine is pinned to the training stack by golden-vector tests spanning the input range, knot boundaries, out-of-domain points, closed-loop states, single cell steps and full rollouts: the floating-point path matches PyTorch within $10^{-4}$ relative error on branch outputs and $5\times10^{-3}$ normalized state error over full rollouts on every exported model, the MLP path is bit-identical to an independent NumPy reference, and the quantized path is pinned within one quantization step against an independent Python implementation. The suite is run under \texttt{qemu-riscv64} and on-device before every hardware session, with integer-path outputs bit-exact between both.

For the dependability evaluation, we implement post-training \emph{static}
quantization in its standard, architecture-agnostic
form~\cite{jacob2018quantization}: symmetric per-tensor weight quantization
($\mathrm{scale}=\max|w|/q_{\max}$) and affine per-tensor activation and
state quantization calibrated from minimum/maximum ranges over the training
trajectories, at INT16, INT8 and, as further degradation points, INT6 and
INT4. Per-tensor scaling is the deliberate choice here rather than the finer
per-channel granularity also in common use, since it is precisely the
granularity the attribution below implicates. No KAN-specific co-design is
applied, so that the unmodified topology is characterized under a scheme
treating both architectures identically, establishing the baseline a
co-designed scheme would have to improve upon~\cite{huang2025hardware}.
Execution follows the standard simulated-integer model: values are held with
intN semantics and dequantized for floating-point arithmetic at tensor
boundaries, the spline recursion's bin search consumes the truncated input, so the
interval-selection pathway specific to spline bases is exercised and can be
isolated, and MLP biases remain unquantized, as int32 biases are in a real
integer pipeline~\cite{jacob2018quantization}. 

This reproduces the \emph{numerical} effect of integer execution exactly but
not its speed, so quantized configurations are evaluated only for accuracy
and stability; no latency or energy claim about integer execution is made.
The memory side of the trade-off is analytic rather than measured: INT8
weight storage is one quarter of FP32 (511\,B--17.8\,KB to approx. 128\,B--4.5\,KB), a reduction Section~\ref{sec:results} shows the KAN cannot rely on, since its
trajectory collapses at INT8 well before the matched MLP's does. Five modes support the attribution analysis of
Section~\ref{sec:results}: \texttt{full}, quantizing state, weights and
activations and requantizing the state every step, as an end-to-end integer
loop would require; \texttt{residual\_only}, a floating-point loop with only
the branch quantized, which removes the state-requantization floor; and isolated single-component modes (\texttt{input}-, \texttt{weights}-, and \texttt{activations}-only).

\section{Experimental Methodology}
\label{sec:methodology}

The eight configurations of Table~\ref{tab:results} are the four KAN and four
MLP architectures evaluated for accuracy in the prior study, so that the
established accuracy transfers unchanged to the hardware measurements. All
eight are trained per system across five random seeds, and the model deployed
is the one achieving the \emph{median} discovery accuracy across those seeds,
not the best, so that the results reflect a representative training outcome. Latency is measured with a single core pinned, $10^4$ warmup calls, and three
interleaved rounds of ten repetitions of $2\times10^4$ calls each over a
fixed buffer of 1024 states drawn from the models' input domain, with a
checksum accumulated over every output to prevent the compiler from
eliminating the computation; we report medians. Both the residual branch
alone and the full HRPINN step are measured, since the branch is what differs
between architectures while the step is what a deployed controller executes.
The CPU frequency governor is pinned to \texttt{performance} throughout,
which removes frequency transitions as a source of run-to-run variance and
ensures that latency and energy are measured at the same operating point. The board is powered through an inline USB coulomb counter at its USB-C
input, so the measurement covers the whole board and each reading is already
a hardware time integral of current, in mAh, with no sampling rate that
could alias the per-step transients. Supply voltage was a stable
$V = 5.11$\,V, so charge converts to energy as $E = Q\,V$ with
$Q\,[\mathrm{C}] = 3.6\,Q\,[\mathrm{mAh}]$. 

An automated script runs a 60\,s idle baseline, then per model a 120\,s
fixed-duration loop of HRPINN steps whose throughput is logged by the same
binary used for latency, separated by 30\,s idle gaps, and a closing idle
baseline; meter readings are timestamped so the idle charge accumulated
during operator readout (at most 21\,s) is subtracted.
Per-step energy is obtained differentially, $E_{\mathrm{step}} = \frac{P_{\mathrm{load}} - P_{\mathrm{idle}}}{\mathrm{steps/s}},$
so that the board's static power is subtracted and only the incremental cost
of running the model is attributed to it. The meter resolves 1\,mAh, which
over a 120\,s window corresponds to 30\,mA, or 153\,mW at 5.11\,V. This
bounds the per-model figures to approximately $\pm$19\%, and bounds any
difference in incremental active power between two architectures to below
153\,mW; the latter is the quantity the energy conclusion rests on. All
energy measurements are FP32. For every model we draw 20 initial conditions uniformly from $[-1.5,1.5]^2$
and run 2000-step (100\,s) closed-loop rollouts at FP32, INT16, INT8, INT6
and INT4, in the quantization modes defined in Section~\ref{sec:framework}.
At every step we record the Euclidean distance, in normalized state space,
between the quantized rollout and the same model's own floating-point rollout
from the same initial condition, which isolates the effect of quantization
from the model's baseline accuracy. We report when this distance first exceeds $0.1$, a threshold safely between floating-point noise and the state-space scale. Because computation is deterministic and bit-reproducible, results are generated on the host and spot-checked on the board.

\section{Results}
\label{sec:results}

\begin{table*}[t]
\begin{minipage}[t]{0.64\textwidth}
\centering
\scriptsize
\scriptsize
\caption{Prior discovery accuracy and measured deployment cost on the
VisionFive~2, for the same trained models. Accuracy is mean $\pm$95\%
bootstrap CI over 100 seeds. Ops: analytical count of scalar floating-point
adds and multiplies per branch inference, excluding the parenthesized
transcendental (\texttt{expf}) calls, each costing several multiply--accumulates.
Latency: median per residual-branch call. Energy: per
HRPINN step, differential, carrying approximately
$\pm$19\% from meter resolution.}
\label{tab:results}
\scriptsize
\setlength{\tabcolsep}{3pt}
\begin{tabular}{llrrrccrr}
\toprule
& & & & & \multicolumn{2}{c}{Discovery $\Rsq$} &
\multicolumn{2}{c}{Energy ($\mu$J)} \\
\cmidrule(lr){6-7}\cmidrule(l){8-9}
Arch. & Config. & Params & Ops (+exp) & Lat. ($\mu$s) & Duffing & VdP & Duffing & VdP \\
\midrule
\multirow{4}{*}{KAN}
 & Very Small & 120  & 1350 (+6)  & 11.1 & $.914{\pm}.061$ & $.743{\pm}.061$ & 3.68  & 4.38 \\
 & Small      & 240  & 2322 (+10) & 18.7 & $.874{\pm}.080$ & $.785{\pm}.073$ & 7.05  & 7.90 \\
 & Wide       & 480  & 4266 (+18) & 33.8 & $.468{\pm}.773$ & $-.602{\pm}2.84$ & 12.97 & 14.52 \\
 & Deep       & 880  & 4986 (+18) & 36.2 & (unstable)      & $.754{\pm}.079$ & 17.74 & 14.70 \\
\midrule
\multirow{4}{*}{MLP}
 & Tiny   & 105  & 209  & 0.83 & $.906{\pm}.092$ & $.622{\pm}.173$ & 0.33  & 0.23 \\
 & Small  & 337  & 673  & 2.33 & $.937{\pm}.047$ & $.879{\pm}.032$ & 1.27  & 1.00 \\
 & Medium & 1185 & 2369 & 7.68 & $.951{\pm}.033$ & $.879{\pm}.019$ & 2.95  & 3.14 \\
 & Large  & 4417 & 8833 & 27.5 & $.932{\pm}.063$ & $.898{\pm}.017$ & 10.56 & 10.98 \\
\bottomrule
\end{tabular}
\end{minipage}\hfill
\begin{minipage}[t]{0.34\textwidth}
\centering
\scriptsize
\caption{Closed-loop divergence under quantization (\texttt{full} mode):
fraction of 20 initial conditions diverged, and median divergence step, out
of a 2000-step (100\,s) horizon. A dash denotes no divergence within the
horizon.}
\label{tab:dependability}
\scriptsize
\setlength{\tabcolsep}{3pt}
\begin{tabular}{llrrrr}
\toprule
& & \multicolumn{2}{c}{INT16} & \multicolumn{2}{c}{INT8} \\
\cmidrule(lr){3-4}\cmidrule(l){5-6}
System & Model & div. & step & div. & step \\
\midrule
\multirow{4}{*}{Duffing}
 & KAN Very Small & 0\,\%   & --- & 100\,\% & \textbf{6.5} \\
 & MLP Tiny       & 0\,\%   & --- & 100\,\% & 282 \\
 & KAN Small      & 0\,\%   & --- & 100\,\% & 119 \\
 & MLP Small      & 0\,\%   & --- & 100\,\% & 316 \\
\midrule
\multirow{4}{*}{Van der Pol}
 & KAN Very Small & 0\,\%   & --- & 100\,\% & 78 \\
 & MLP Tiny       & 0\,\%   & --- & 100\,\% & 196 \\
 & KAN Small      & 100\,\% & 575 & 100\,\% & 93 \\
 & MLP Small      & 0\,\%   & --- & 100\,\% & 146 \\
\bottomrule
\end{tabular}
\end{minipage}
\vspace{-5mm}
\end{table*}

\textbf{H1 is confirmed.} The size tiers pair KAN Very Small with MLP Tiny, KAN Small
and KAN Wide with MLP Small, and KAN Deep with MLP Medium. Only the first two
are also accuracy-comparable in Table~\ref{tab:results}: KAN Wide and KAN
Deep are the configurations the prior study found unstable, so their tiers
test scaling rather than equal-accuracy cost. For the two
accuracy-comparable pairs, KAN Very Small (120 params) takes 11.1\,$\mu$s per
call against 0.83\,$\mu$s for MLP Tiny (105 params), a $13.5\times$
difference, and KAN Small is $8.0\times$ slower than MLP Small; across all
four tiers the ratio spans $4.7\times$--$14.5\times$. Normalized per
parameter, the KAN spends 41--93\,ns (mean 73) against 6.2--8.0\,ns for the
MLP (mean 6.9), an order of magnitude apart and the reverse of what parameter
counts alone would suggest: MLP latency tracks its analytical operation count
closely across the full range, while the KAN's latency per operation is
consistently higher. The optimized reciprocal variant recovers 6--18\% of KAN latency without altering the trend. Both system instances (Duffing and Van der Pol, with different knot vectors) agree within 1.5\%, showing KAN latency on this core is boundable from architecture alone, as worst-case execution-time analysis requires.

\textbf{H2 is also confirmed.} Over the fixed 120\,s window every model drew 8--10\,mAh, constant to within the meter's 1\,mAh resolution, while the throughput of those same runs varied by a factor of 40 ($2.8\times10^4$ to $1.15\times10^6$ steps/s): the board draws essentially the same power whatever it computes, and what differs is how many integration steps that power buys. Two derived quantities confirm this: the incremental active power $P_{\mathrm{load}}-P_{\mathrm{idle}}$ spans 268--537\,mW, only 1.75 meter counts and so indistinguishable from a constant, and the energy-ratio to latency-ratio quotient over the four tiers on both systems has mean 1.01 with scatter 0.70--1.46, consistent with the $\pm$27\% a propagated half-count error produces. The KAN's energy penalty therefore equals rather than compounds its latency penalty: the accuracy-comparable pairs cost $11.3\times$ and $5.6\times$ more energy, and all four tiers span $4.7\times$--$18.7\times$.

\textbf{Finally, Table~\ref{tab:dependability} confirms H3.} At INT16 every pair stays within the
threshold for the full 100\,s horizon, the sole exception being a KAN
configuration already flagged as unstable in the prior study. At INT8 in
\texttt{full} mode, where the recurrent state itself is requantized every
step, \emph{every} model eventually crosses the threshold, because an 8-bit
state resolution ($2/255$ of the normalized range) compounds over a long
enough rollout regardless of architecture; the architectures differ sharply
in \emph{when}. Duffing KAN Very Small diverges at a median of 6.5 steps
(0.3\,s) against 282 steps (14\,s) for the matched MLP, a $43\times$
difference, with the same ordering in every pair. The effect is one of
precision and not of measurement: the non-diverging fraction decreases
monotonically from FP32 through INT16 to INT8, INT6 and INT4 for all models,
and the only models diverging already at INT16 are those the prior study
independently flagged as unstable. Under \texttt{residual\_only}, which
removes the state-requantization floor, the gap persists: on Duffing, KAN
Small diverges at a median step of 137 against 1183 for MLP Small.
Quantizing one component at a time isolates the responsible mechanism.
Quantizing only the input, a pathway with no analogue in a dense layer,
since a perturbed input can select a different set of basis functions rather
than merely perturb a product, is essentially harmless to both architectures
($3\times10^{-6}$ and $2\times10^{-6}$ median trajectory MSE against each
model's own FP32 rollout): at INT8 the input quantization step
($\approx0.008$) is about $50\times$ smaller than a spline grid interval
($0.4$), so it rarely moves an input into a neighbouring interval. Quantizing
only the weights reproduces most of the full-loop degradation for the KAN
($0.204$ against $0.275$) but almost none of it for the MLP
($6\times10^{-5}$ against $4\times10^{-3}$), with activations-only
intermediate for the KAN ($0.048$) and negligible for the MLP
($2\times10^{-5}$). Recomputing discovery $\Rsq$ through the quantized engine ties this to the
accuracy metric: on Duffing, KAN Very Small falls from $0.90$ to $-0.10$ at
INT8 and KAN Deep to $-0.20$, while MLP Tiny and MLP Medium retain $0.92$
and $0.94$; on Van der Pol both degrade by $\leq0.02$, so the collapse
depends on the trained weight distribution.

\section{Discussion and Conclusion}
\label{sec:discussion}

Three questions follow from these measurements: why energy tracks latency so
closely, whether kernel engineering could close the latency gap, and what the
quantization attribution implies for KAN co-design. The first is a property
of the operating regime. With the governor pinned the core runs at a fixed
clock, and the difference in incremental active power between a spline
recursion and a matrix-vector product is below the meter's resolution, so
energy per step is constant power multiplied by execution time. That proxy
would not survive an on-demand governor, under which a longer-running
workload also tends to drive the core to a higher operating point, and the
energy penalty could then exceed the latency penalty. The second question has a quantitative answer: two classical embedded-ML optimizations bound how far a tuned scalar kernel could go. Replacing the SiLU base path's
\texttt{expf} with a lookup table removes at most the 6--18 transcendental
calls per inference of Table~\ref{tab:results}, against 1350--4986 scalar
operations, so at a generous twenty operations per call it addresses under
10\% of the work. Table-driven B-spline evaluation targets the dominant term
instead, but training applies grid updates, so every edge carries its own
non-uniform knot vector and the table would have to be materialized per
edge, trading away the parameter-count advantage that motivates KANs in the
first place; dedicated designs likewise find lookup-table mappings of
B-splines circuit-expensive~\cite{huang2025hardware}. Our
precomputed-reciprocal variant, which eliminates every division from the
recursion, recovers 6--18\%. Together these bound arithmetic-level tuning at
roughly a quarter of the KAN's cost, against gaps of $8.0\times$ and
$13.5\times$ on the accuracy-comparable pairs: a fully optimized scalar
kernel would remain an order of magnitude behind.

The attribution result rules out a failure mode with no dense-layer
analogue, input misassignment across knot intervals, and concentrates the
damage in a mechanism both architectures share, weight quantization, where
what differs is the coefficient distribution rather than the operation. This extends to a
closed recurrent loop, and to trajectory divergence rather than
classification accuracy, what component-wise studies of feedforward KANs
report: that the learnable coefficients are the most quantization-sensitive
tensor and their heterogeneous distribution is what makes low-precision KAN
deployment hard~\cite{errabii2026kantize}. A plausible
structural reason is that a KAN layer's spline-coefficient tensor
holds sub-functions with heterogeneous per-edge magnitudes, so a single
per-tensor scale compresses the small coefficients shaping individual spline
segments far more than it compresses an MLP's comparatively homogeneous
weight matrix. If this generalizes, per-edge or per-channel spline-weight
scales are a more promising co-design direction than finer input grids. What
such co-design must overcome is quantified by the contrast with dedicated
accelerators, which report order-of-magnitude energy and area gains over a
DNN baseline~\cite{huang2025hardware}: without co-design, on a
general-purpose scalar core, the same formulation costs
$4.7\times$--$14.5\times$ more and fails at INT8 where the MLP survives. An
MLP residual branch is therefore the more dependable default in the regimes
evaluated, while a vanilla KAN remains defensible where memory footprint is
the binding constraint, the residual is additively separable, and execution
stays at FP32 or INT16.

These conclusions hold within a deliberately narrow scope. The two oscillators were chosen so that every hardware measurement could be
paired with an exactly computable discovery accuracy, which an unidentified
real plant does not permit without introducing identification error as a
confound. They are not a complete verification set: both are second-order,
two-state, autonomous and unforced, and they do not exercise higher state
dimensionality, external forcing, discontinuous or hysteretic dynamics,
stiff systems requiring implicit integration, or multi-rate behavior. What
they do span is the structural axis this comparison turns on, an additively
separable residual against a multiplicatively coupled one; and since the
per-step cost of both architectures is an exactly countable function of
layer width and depth rather than of state dimensionality, the cost
mechanisms carry over directionally to larger plants built from the same
primitives, though accuracy conclusions would need re-establishing per
system. 

On the measurement side, the meter's 1\,mAh resolution bounds any
difference in incremental active power between the architectures at
153\,mW, and it is that bound, not any per-model figure, that supports the
energy conclusion; five board sessions differed by less than instrument
resolution, so one representative session is reported, and finer attribution
would require core-rail instrumentation. Quantized execution is simulated at
the arithmetic level rather than run through a genuine integer pipeline, so
the precision findings support accuracy and stability conclusions but not
integer latency or energy claims. Finally, the evaluation covers one scalar
core and standard per-tensor post-training quantization, and a
vector-capable core would change both architectures' constants, though the
KAN's gather-heavy, branchy recursion is in principle harder to vectorize
than dense matrix products.

Taken together, these measurements indicate that the parameter efficiency
reported for KANs on separable residuals does not transfer to deployment
cost on a scalar embedded core, and that neither classical kernel tuning nor
architecture-agnostic post-training quantization closes the gap. Future work
should evaluate co-designed quantization and per-edge tabulated basis
evaluation, including the memory-for-latency exchange the latter implies,
against the baseline established here; extend the methodology to
higher-dimensional and eventually industrial plants now that the
architectural variables are characterized in isolation; and implement both
architectures with RISC-V vector extensions to test whether the scalar gap
narrows or widens under data-level parallelism.


\balance
\bibliographystyle{IEEEtran}
\bibliography{references}

@inproceedings{liu2024kan,
 author = {Liu, Ziming and Wang, Yixuan and Vaidya, Sachin and Ruehle, Fabian and Halverson, James and Soljacic, Marin and Hou, Thomas and Tegmark, Max },
 booktitle = {International Conference on Learning Representations},
 editor = {Y. Yue and A. Garg and N. Peng and F. Sha and R. Yu},
 pages = {70367--70413},
 title = {KAN: Kolmogorov\textendash Arnold Networks},
 url = {https://proceedings.iclr.cc/paper_files/paper/2025/file/afaed89642ea100935e39d39a4da602c-Paper-Conference.pdf},
 volume = {2025},
 year = {2025}
}

@article{koenig2024kan,
  title={KAN-ODEs: Kolmogorov--Arnold network ordinary differential equations for learning dynamical systems and hidden physics},
  author={Koenig, Benjamin C. and Kim, Suyong and Deng, Sili},
  journal={Computer Methods in Applied Mechanics and Engineering},
  volume={432},
  pages={117397},
  year={2024},
  publisher={Elsevier},
  doi={10.1016/j.cma.2024.117397}
}

@inproceedings{huang2025hardware,
  title={Hardware Acceleration of Kolmogorov-Arnold Network (KAN) for Lightweight Edge Inference},
  author={Huang, Wei-Hsing and Jia, Jianwei and Kong, Yuyao and Waqar, Faaiq and Wen, Tai-Hao and Chang, Meng-Fan and Yu, Shimeng},
  booktitle={Proceedings of the 30th Asia and South Pacific Design Automation Conference (ASP-DAC)},
  pages={693--699},
  year={2025},
  publisher={ACM},
  doi={10.1145/3658617.3697677},
}

@article{khalid2026hardware,
  title={Hardware-Oriented Inference Complexity of Kolmogorov-Arnold Networks},
  author={Khalid, Bilal and Freire, Pedro J. and Turitsyn, Sergei K. and Prilepsky, Jaroslaw E.},
  journal={IEEE Access},
  year={2026},
  doi={10.1109/ACCESS.2026.3707241},
  note={Also available at: https://arxiv.org/abs/2604.03345}
}

@article{brunke2022safe,
  title={Safe Learning in Robotics: From Learning-Based Control to Safe Reinforcement Learning},
  author={Brunke, Lukas and Greeff, Melissa and Hall, Adam W. and Yuan, Zhaocong and Zhou, Siqi and Panerati, Jacopo and Schoellig, Angela P.},
  journal={Annual Review of Control, Robotics, and Autonomous Systems},
  volume={5},
  pages={411--444},
  year={2022},
  doi={10.1146/annurev-control-042920-020211}
}

@article{chen2024hard,
  title={Physics-Informed Neural Networks with Hard Linear Equality Constraints},
  author={Chen, Hao and Constante Flores, Gonzalo E. and Li, Can},
  journal={Computers \& Chemical Engineering},
  volume={189},
  pages={108764},
  year={2024},
  publisher={Elsevier},
  doi={10.1016/j.compchemeng.2024.108764},
}

@article{raissi2019pinn,
  title={Physics-informed neural networks: A deep learning framework for solving forward and inverse problems involving nonlinear partial differential equations},
  author={Raissi, Maziar and Perdikaris, Paris and Karniadakis, George E.},
  journal={Journal of Computational Physics},
  volume={378},
  pages={686--707},
  year={2019},
  publisher={Elsevier},
  doi={10.1016/j.jcp.2018.10.045}
}

@article{noorizadegan2025practitioner,
  title={A Practitioner's Guide to Kolmogorov-Arnold Networks},
  author={Noorizadegan, Amir and Wang, Sifan and Ling, Leevan and Dominguez-Morales, Juan P.},
  journal={Computer Science Review},
  volume={62},
  pages={100991},
  year={2026},
  publisher={Elsevier},
  doi={10.1016/j.cosrev.2026.100991}
}

@article{spotorno2025hard,
  title={Hard-Constrained Neural Networks with Physics-Embedded Architecture for Residual Dynamics Learning and Invariant Enforcement in Cyber-Physical Systems},
  author={Spotorno, Enzo Nicol{\'a}s and Leal Filho, Josafat and Fr{\"o}hlich, Ant{\^o}nio Augusto},
  journal={arXiv preprint arXiv:2511.23307},
  year={2025},
  eprint={2511.23307},
  archivePrefix={arXiv},
  primaryClass={cs.LG}
}

@misc{spotorno2026kaniclr,
  title={Empirical Stability Analysis of Kolmogorov-Arnold Networks in Hard-Constrained Recurrent Physics-Informed Discovery},
  author={Spotorno, Enzo Nicol{\'a}s and Leal Filho, Josafat and Fr{\"o}hlich, Ant{\^o}nio Augusto},
  year={2026},
  eprint={2602.09988},
  archivePrefix={arXiv},
  primaryClass={cs.LG},
  note={Poster at the AI\&PDE Workshop, ICLR 2026}
}

@article{liu2025structured,
  title={Structured Kolmogorov-Arnold Neural ODEs for Interpretable Learning and Symbolic Discovery of Nonlinear Dynamics},
  author={Liu, Wei and Bacsa, Kiran and Tang, Loon Ching and Chatzi, Eleni},
  journal={arXiv preprint arXiv:2506.18339},
  year={2025},
  eprint={2506.18339},
  archivePrefix={arXiv},
  primaryClass={cs.LG}
}

@article{mostajeran2025scaled,
  title={Scaled-cPIKANs: Spatial variable and residual scaling in Chebyshev-based physics-informed Kolmogorov-Arnold networks},
  author={Mostajeran, Farinaz and Faroughi, Salah A.},
  journal={Journal of Computational Physics},
  volume={537},
  pages={114116},
  year={2025},
  publisher={Elsevier},
  doi={10.1016/j.jcp.2025.114116}
}

@article{abueidda2025deepokan,
  title={DeepOKAN: Deep operator network based on Kolmogorov Arnold networks for mechanics problems},
  author={Abueidda, Diab W. and Pantidis, Panos and Mobasher, Mostafa E.},
  journal={Computer Methods in Applied Mechanics and Engineering},
  volume={436},
  pages={117699},
  year={2025},
  publisher={Elsevier},
  doi={10.1016/j.cma.2024.117699}
}

@article{zhang2025physics,
  title={Physics-informed neural networks with hybrid Kolmogorov-Arnold network and augmented Lagrangian function for solving partial differential equations},
  author={Zhang, Zhaoyang and Wang, Qingwang and Zhang, Yinxing and Shen, Tao and Zhang, Weiyi},
  journal={Scientific Reports},
  volume={15},
  number={1},
  pages={10523},
  year={2025},
  publisher={Nature Publishing Group UK London},
  doi={10.1038/s41598-025-92900-1}
}

@misc{visionfive2,
  title={VisionFive 2 Datasheet},
  author={{StarFive Technology}},
  year={2023},
  note={Version 1.53, accessed 27 July 2026},
  url={https://doc-en.rvspace.org/VisionFive2/Datasheet/VisionFive_2/introduction_ds.html}
}

@article{nascimento2021hybrid,
  title={Hybrid physics-informed neural networks for lithium-ion battery modeling and prognosis},
  author={Nascimento, Renato G. and Corbetta, Matteo and Kulkarni, Chetan S. and Viana, Felipe A. C.},
  journal={Journal of Power Sources},
  volume={513},
  pages={230526},
  year={2021},
  publisher={Elsevier},
  doi={10.1016/j.jpowsour.2021.230526}
}

@article{rai2020hybrid,
  title={Driven by Data or Derived Through Physics? A Review of Hybrid Physics Guided Machine Learning Techniques With Cyber-Physical System (CPS) Focus},
  author={Rai, Rahul and Sahu, Chandan K.},
  journal={IEEE Access},
  volume={8},
  pages={71050--71073},
  year={2020},
  publisher={IEEE},
  doi={10.1109/ACCESS.2020.2987324}
}

@article{shukla2024fair,
title = {A comprehensive and FAIR comparison between MLP and KAN representations for differential equations and operator networks},
journal = {Computer Methods in Applied Mechanics and Engineering},
volume = {431},
pages = {117290},
year = {2024},
issn = {0045-7825},
doi = {https://doi.org/10.1016/j.cma.2024.117290},
url = {https://www.sciencedirect.com/science/article/pii/S0045782524005462},
author = {Khemraj Shukla and Juan Diego Toscano and Zhicheng Wang and Zongren Zou and George Em Karniadakis},
}

@inproceedings{jacob2018quantization,
  title={Quantization and Training of Neural Networks for Efficient
         Integer-Arithmetic-Only Inference},
  author={Jacob, Benoit and Kligys, Skirmantas and Chen, Bo and Zhu, Menglong
          and Tang, Matthew and Howard, Andrew and Adam, Hartwig and
          Kalenichenko, Dmitry},
  booktitle={Proc. IEEE/CVF Conf. on Computer Vision and Pattern Recognition (CVPR)},
  pages={2704--2713},
  year={2018},
  doi={10.1109/CVPR.2018.00286}
}

@article{errabii2026kantize,
  title={{KANtize}: Exploring Low-bit Quantization of Kolmogorov-Arnold
         Networks for Efficient Inference},
  author={Errabii, Sohaib and Sentieys, Olivier and Traiola, Marcello},
  journal={arXiv preprint arXiv:2603.17230},
  year={2026}
}

\end{document}